%% file: main.tex
\documentclass[10pt,twocolumn,letterpaper]{article}

\usepackage[pagenumbers]{cvpr} 

\input{preamble}

\definecolor{cvprblue}{rgb}{0.21,0.49,0.74}
\usepackage[pagebackref,breaklinks,colorlinks,allcolors=cvprblue]{hyperref}

\def\paperID{XXX} 
\def\confName{CVPR}
\def\confYear{2025}

\title{SplatAD: Real-Time Lidar and Camera Rendering with 3D Gaussian Splatting for Autonomous Driving}

\author{Georg Hess$^{\dagger,1,2}$
\quad Carl Lindström$^{\dagger,1,2}$
\quad Maryam Fatemi$^{1}$
\quad Christoffer Petersson$^{1,2}$
\quad Lennart Svensson$^{2}$
\\
\normalsize$^1$Zenseact \hspace{0.8cm} $^2$Chalmers University of Technology \hspace{0.8cm}\\
{\tt\small \{firstname.lastname\}@\{zenseact.com, chalmers.se\}}
}

\begin{document}

\twocolumn[{
\maketitle
\begin{center}
    \captionsetup{type=figure}
    \includegraphics[width=1.0\textwidth,trim={0cm 2.8cm 0cm 0cm},clip]{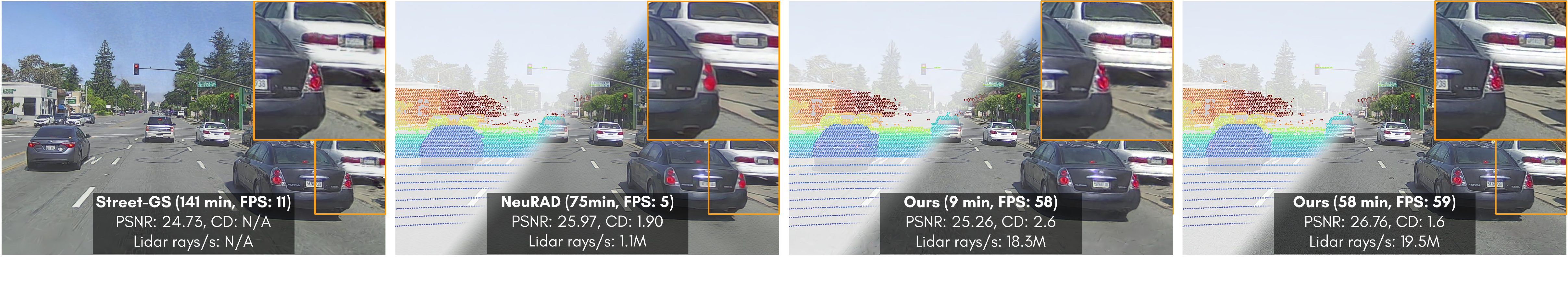}
    \captionof{figure}{{\modelname} is the first method capable of realistic camera and lidar rendering using 3D Gaussian Splatting. Whereas previous methods are either fast or multi-modal, {\modelname} enables real-time, high-quality rendering for both camera and lidar. In addition, {\modelname} can reach competitive performance, \eg, PSNR for image and Chamfer distance for point cloud, within minutes.}
    \label{fig:frontpage}
\end{center}
}]

\input{sec/0_abstract}
\def\thefootnote{$\dagger$}\footnotetext{\vspace{-3mm}These authors contributed equally to this work.}\def\thefootnote{\arabic{footnote}}
\input{sec/1_intro}
\input{sec/2_relatedwork}
\input{sec/3_method}

\input{sec/4_experiments}

\input{sec/5_conclusion}
{
    \small
    \bibliographystyle{ieeenat_fullname}
    \bibliography{main}
}

\input{sec/6_suppl}

\end{document}

%% file: preamble.tex
\DeclareMathOperator{\arctantwo}{arctan2}
\def\modelname{{SplatAD}}

\newcommand{\parsection}[1]{\noindent\textbf{#1:}}
\usepackage{multirow}

\def\bbf{{\bf b}}

\def\dbf{{\bf d}}
\def\ebf{{\bf e}}
\def\fbf{{\bf f}}
\def\Fbf{{\bf F}}
\def\Ibf{{\bf I}}
\def\Jbf{{\bf J}}
\def\obf{{\bf o}}
\def\pbf{{\bf p}}
\def\Sbf{{\bf S}}
\def\qbf{{\bf q}}
\def\vbf{{\bf v}}
\def\xbf{{\bf x}}

\def\mubf{{\boldsymbol{\mu}}}
\def\Sigmabf{{\boldsymbol{\Sigma}}}
\def\omegabf{{\boldsymbol{\omega}}}
\def\Deltabf{{\boldsymbol{\Delta}}}

\usepackage{tabulary}
\usepackage{colortbl}
\definecolor{tabfirst}{rgb}{1, 0.7, 0.7} 
\definecolor{tabsecond}{rgb}{1, 0.85, 0.7} 
\definecolor{tabthird}{rgb}{1, 1, 0.7} 

\usepackage{gensymb}

\usepackage{mathtools}
\DeclarePairedDelimiter\ceil{\lceil}{\rceil}
\DeclarePairedDelimiter\floor{\lfloor}{\rfloor}

\newcommand{\Mod}[1]{\ (\mathrm{mod}\ #1)}

%% file: sec/0_abstract.tex
\begin{abstract}
Ensuring the safety of autonomous robots, such as self-driving vehicles, requires extensive testing across diverse driving scenarios. Simulation is a key ingredient for conducting such testing in a cost-effective and scalable way. Neural rendering methods have gained popularity, as they can build simulation environments from collected logs in a data-driven manner. However, existing neural radiance field (NeRF) methods for sensor-realistic rendering of camera and lidar data suffer from low rendering speeds, limiting their applicability for large-scale testing. While 3D Gaussian Splatting (3DGS) enables real-time rendering, current methods are limited to camera data and are unable to render lidar data essential for autonomous driving. To address these limitations, we propose {\modelname}, the first 3DGS-based method for realistic, real-time rendering of dynamic scenes for both camera and lidar data. {\modelname} accurately models key sensor-specific phenomena such as rolling shutter effects, lidar intensity, and lidar ray dropouts, using purpose-built algorithms to optimize rendering efficiency. Evaluation across three autonomous driving datasets demonstrates that {\modelname} achieves state-of-the-art rendering quality with up to +2 PSNR for NVS and +3 PSNR for reconstruction while increasing rendering speed over NeRF-based methods by an order of magnitude.
See \href{https://research.zenseact.com/publications/splatad/}{here} for our project page.
\end{abstract}

%% file: sec/1_intro.tex
\section{Introduction}

Large-scale testing is essential to ensure the safety of autonomous robots, such as self-driving vehicles (SDVs), before real-world deployment. Data-driven methods that generate digital twins from collected logs offer a scalable way to build diverse, realistic simulation environments for testing. Unlike real-world testing, which can be costly, time-intensive, and limited by physical constraints, simulation enables rapid, low-cost exploration of multiple scenarios, helping to optimize SDVs for safety, comfort, and efficiency. Motivated by this, a multitude of methods have emerged based on neural radiance fields (NeRFs)~\cite{mildenhall2021nerf,unisim,tonderski2024neurad,turki2023suds} and 3D Gaussian Splatting (3DGS)~\cite{kerbl20233d,zhou2024drivinggaussian,yan2024street}.

Recent NeRF-based methods~\cite{tonderski2024neurad,unisim} provide high-fidelity sensor simulation jointly for both camera and lidar, matching the most common sensor rigs in popular autonomous driving (AD) datasets~\cite{caesar2020nuscenes,alibeigi2023zenseact,Argoverse2}. However, the slow rendering speed of NeRF-based methods makes them costly and challenging to use for large-scale testing. 3DGS provides an attractive alternative to NeRFs, as its accelerated rendering achieves comparable image realism while increasing inference speeds by an order of magnitude. Nevertheless, 3DGS-based methods for the AD setting~\cite{zhou2024drivinggaussian,yan2024street,chen2024omnire} inherit the limitation of only being able to render camera data, overlooking the lidar modality. The lidar's ability to directly perceive the 3D environment makes it a powerful tool in modern AD stacks and, consequently, an important modality to simulate.

In this paper, we aim to solve efficient, differentiable, and realistic rendering of camera \textit{and} lidar data using 3D Gaussian Splatting. Applying 3DGS to lidar sensors presents unique challenges due to their distinct characteristics. First, unlike cameras, which capture dense, regularly spaced pixels, lidars record sparse, non-linear point clouds with large empty gaps from non-returning rays. Second, most lidars capture a full 360° view of the scene, where existing methods often project data into multiple depth images~\cite{chen2024omnire}—an inefficient approach that overlooks the sparse structure of the lidar. Last, lidars suffer from a rolling shutter effect, with each sweep taking up to 100 ms, during which the ego vehicle may move several meters, violating 3DGS’s single-origin assumption.
To overcome these challenges, we introduce {\modelname}, a novel view synthesis method that unifies camera and lidar rendering and is designed for real-time rendering of large-scale dynamic traffic scenes. Our method replaces spherical harmonics with per-Gaussian learnable features to jointly model sensor-specific phenomena\textemdash from lidar ray drop and intensity variations to camera-specific appearance changes. By introducing purpose-built rasterization algorithms for rendering lidar in spherical coordinates, we achieve both superior efficiency and realism. Further, we demonstrate efficient methods for modeling rolling shutter effects across both sensor modalities. We verify the effectiveness and generalizability of our method across three popular automotive datasets, achieving state-of-the-art results on all of them.
To summarize, our contributions are as follows:
\begin{itemize}
    \item We propose the first method for efficient lidar rendering using 3D Gaussians, introducing custom CUDA-accelerated algorithms for rasterizing sparse point clouds in spherical coordinates.
    \item We introduce the first 3DGS method capable of rendering both camera and lidar from a unified representation, enabling accelerated scaling of novel view synthesis for automotive applications.
    \item We present effective techniques for realistic sensor modeling using 3D Gaussians, enabling accurate handling of rolling shutter, lidar intensity, ray dropping, and variations in sensor appearance.
    \item Through extensive evaluation on three popular automotive datasets, we demonstrate state-of-the-art results across all benchmarks, validating our method's effectiveness and generalizability.
\end{itemize}

%% file: sec/2_relatedwork.tex
\section{Related work}
\label{sec:related_work}
\parsection{NeRFs for automotive data}
Since the introduction of NeRFs, neural representations have become central to 3D reconstruction and novel view synthesis \cite{mildenhall2021nerf, barron2021mip, barron2022mip360, Barron_2023_ICCV}. 
Many works have applied NeRF-based methods to automotive data \cite{unisim, tonderski2024neurad, turki2023suds, yang2024emernerf}, enabling sensor-realistic renderings of novel views in large and dynamic scenes. Recent advances provide realism sufficient for downstream applications \cite{lindstrom2024nerfs, ljungbergh2025neuroncap}.
While early methods focused on either camera \cite{ost2021neural, xie2023snerf, kundu2022panoptic, fu2022panoptic} or lidar \cite{huang2023neural,wu2024dynamic, zheng2024lidar4d}, newer methods aim to handle them jointly.
UniSim \cite{unisim} showcased realistic renderings for front camera and 360{\degree} lidar on data from PandaSet \cite{pandaset}. 
The method uses a hash grid representation \cite{muller2022instant}, with separate features for the sky, static background, and dynamic actors, and decodes the color and lidar intensity from volume-rendered features.
NeuRAD \cite{tonderski2024neurad} proposes a simplified network architecture and improved sensor modeling, achieving state-of-the-art results for the full 360\degree~ camera and lidar rig. 
Even so, the slow rendering speeds of NeRF-based methods make scaling costly and challenging. 
{\modelname} aims to overcome this limitation by using rasterization-based CUDA-accelerated algorithms to render both camera and lidar data while, inspired by NeuRAD, emphasizing the modeling of important sensor characteristics for increased realism.

\parsection{3DGS for automotive data}
3DGS \cite{kerbl20233d} uses an explicit representation paired with rasterization techniques and custom hardware-accelerated algorithms to achieve real-time rendering.
Additionally, 3DGS has been shown to create faithful scene reconstructions with rendering quality close to recent NeRF-based methods. 
As a result, several works have applied 3DGS for camera rendering on automotive data \cite{yan2024street, chen2024omnire, khan2024autosplat, zhou2024drivinggaussian, Zhou_2024_CVPR}. 
Periodic Vibration Gaussian (PVG)~\cite{chen2023periodic} applies 3DGS to dynamic scenes by learning the 3D flow of all Gaussians. 
Yet, PVG's lack of an explicit actor representation limits the method's controllability and, hence, applicability for simulation. 
Street Gaussians~\cite{yan2024street}, like Unisim and NeuRAD, instead decomposes the scene into static background and rigid dynamic actors using 3D bounding boxes. 
The authors further extend 3DGS by adding Fourier coefficients to capture time-varying characteristics of dynamic actors. 
Moreover, to handle the sky regions, their method uses semantic masks and a view-dependent cube map. 
OmniRe \cite{chen2024omnire} enhances the scene graph composition with non-rigid nodes for better modeling of pedestrians and cyclists. 
However, this requires pre-processing to track and align human body poses~\cite{goel2023humans} for initializing SMPL~\cite{smpl} pose parameters.

PVG, Street Gaussians, and OmniRe all use lidar points for initialization and depth supervision. However, their point clouds are generated by projecting lidar points into depth images, which is both inefficient and not applicable in novel views. As a consequence, neither of these methods can model important lidar characteristics such as intensity variations, ray drop, and rolling shutter. In contrast, our method supports efficient rendering of both camera and lidar from 3D Gaussians in novel views, and models important sensor characteristics for both modalities.

%% file: sec/3_method.tex
\section{Method}
\label{sec:method}
\begin{figure*}[t]
    \centering
    \includegraphics[width=1\linewidth,trim={0cm 15.95cm 0cm 0cm},clip]{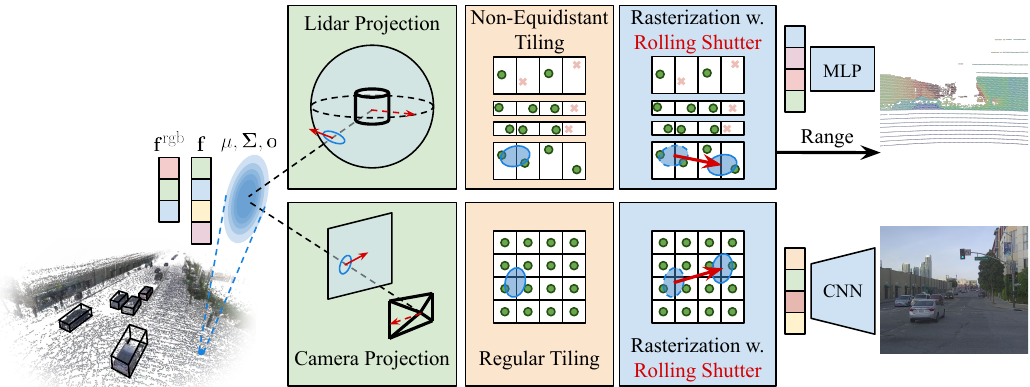}
    \caption{Overview of our proposed method. Given the composition of static and dynamic 3D Gaussians, {\modelname} is capable of differentiable rendering of both lidar and camera data. Our proposed lidar rendering matches the image rendering on a high level, but modifies each component to accurately model sensor characteristics. Our method projects 3D Gaussians with associated feature vectors onto the corresponding sensor modalities (camera and lidar) and employs sensor-specific tiling to match their distinct characteristics. During rasterization, the projected Gaussians are corrected for rolling shutter effects caused by the movement of sensors and, potentially, their own velocity. Finally, the rasterized features are decoded into the respective image and lidar point cloud representations.}
    \label{fig:method_overview}
\end{figure*}
Our aim is to learn a scene representation from collected vehicle logs that enables rendering of realistic camera and lidar data, with the ability to alter the locations of both the ego vehicle and other actors. To scale effectively, the rendering process must be fast, as faster inference speeds enhance its practicality for applications. In the following, we describe our scene representation (\cref{sec:scene_representation}), rendering algorithms (\cref{sec:camera_rendering,sec:lidar_rendering}), and implementation and optimization strategy (\cref{sec:optimization}), see \cref{fig:method_overview} for an overview.

\subsection{Scene representation}
\label{sec:scene_representation}
Our scene representation builds upon 3DGS~\cite{kerbl20233d}, but with key changes to handle the specifics of AD scenes and to enable both camera and lidar rendering from the same representation.
Like 3DGS, each scene is represented by a set of translucent 3D Gaussians with learnable occupancy $o\in(0,1)$, mean $\mathbf{\mubf}\in\mathbb{R}^3$, and covariance matrix ${\Sigmabf}\in\mathbb{R}^{3\times3}$, which in turn is parameterized by scale $\Sbf\in\mathbb{R}^3$ and quaternion $\qbf\in\mathbb{R}^4$ parameters. 
Instead of spherical harmonics~\cite{kerbl20233d}, we assign each Gaussian a learnable base color $\fbf^\text{rgb}\in\mathbb{R}^3$ and feature vector $\fbf\in\mathbb{R}^{D_\fbf}$, where $\fbf$ is used for both view-dependent effects and lidar properties. 
Last, our representation contains a learnable embedding per sensor to model their specific appearance characteristics.

To handle dynamics, we follow the commonly used scene graph decomposition \cite{ost2021neural,yan2024street,zhou2024drivinggaussian,tonderski2024neurad} and divide the scene into a static background and a set of dynamic actors. Each dynamic actor is described by a 3D bounding box and a sequence of SE(3) poses, obtained either from an off-the-shelf object detector and tracker, or annotations. Each Gaussian has a non-learnable ID, indicating whether it is assigned to the static world, or to which actor it belongs. For Gaussians assigned to actors, their mean and covariance are expressed in the local coordinate system of the corresponding axis-aligned bounding box. To compose the scene at a given time $t$, Gaussians assigned to bounding boxes are transformed to world coordinates based on their actor's pose. Since pose estimates for actors may contain inaccuracies, we adjust these with learnable offsets. Furthermore, each actor has a velocity initialized from pose differences and a learnable velocity offset.

\subsection{Camera rendering}
\label{sec:camera_rendering}
Given a posed camera, we compose the set of Gaussians at the corresponding capture time $t$ and render an image $I$ using the efficient tile-based rendering from 3DGS~\cite{kerbl20233d}. While we retain 3DGS's high-level steps---projection and view frustum culling, tile-assignment, depth sorting, and tile-based rasterization---we introduce key adaptations to better model the unique characteristics of AD data. 

\parsection{Projection, tiling, and sorting} Each mean and covariance is transformed from world to camera coordinates, giving $\mubf^C$ and $\Sigmabf^C$. The means are then converted to image space $\mubf^I\in\mathbb{R}^2$ using the perspective projection, while the covariances are transformed using the two upper rows of the Jacobian of the projection, $\Sigmabf^I=\Jbf^I\Sigmabf^C(\Jbf^I)^\text{T} \in\mathbb{R}^{2\times2}$. Gaussians outside the view frustum are culled, where, for efficiency, 3DGS approximates the extent of the Gaussians using a square axis-aligned bounding box (AABB) that covers their 99\% confidence level. Further, 3DGS divides the image into tiles, each measuring $16\times16$ pixels, and assigns Gaussians to all tiles they intersect, duplicating them if necessary. This way, each pixel only has to process a subset of all Gaussians during rasterization. Lastly, the Gaussians are sorted based on the $z$-depth of their mean $\mubf^C$. 

\parsection{Rolling shutter} Many cameras use a rolling shutter, where the capture of an image is not instantaneous, but is done row-by-row, allowing the camera to move during the exposure. Previous work \cite{tonderski2024neurad} has highlighted the importance of modeling rolling shutter effects due to the high sensor velocities present in AD data, where their ray-tracing-based method can easily account for this by shifting the origins of each ray. For 3DGS, the equivalent would require projecting all 3D Gaussians to all camera poses encountered during an exposure, as the Gaussians' positions relative to the camera change with time. As this would be prohibitively expensive, we draw inspiration from \cite{seiskari2024gaussian} and instead approximate the effects directly in 2D image space. Specifically, each Gaussian's velocity relative to the camera is projected to image space, and their pixel mean $\mubf^I$ is adjusted during rasterization based on the capture time of the pixel. However, \cite{seiskari2024gaussian} considered static scenes only, hence we adjust the formulation to account for the dynamics as well.

For each Gaussian, its pixel velocity is approximated as
\begin{equation}
\label{eq:pixel_velocity}
    \vbf^I = \Jbf^I (-\omegabf_C \times \mubf^C - \vbf_C + \vbf_{\text{dyn}}) \in \mathbb{R}^2,
\end{equation}
where $\omegabf_C$ and $\vbf_C$ denote the angular and linear velocity of the camera expressed in the camera coordinate system, and
\begin{equation}
    \vbf_{\text{dyn}} = T^{\text{act}\rightarrow C}(\omegabf_\text{act} \times \mubf^\text{act} + \vbf_\text{act}) \in \mathbb{R}^3,
\end{equation}
denotes the velocity of the Gaussian induced by the angular $\omegabf^\text{dyn}$ and linear velocity $\vbf^\text{dyn}$ of its associated actor (expressed in actor coordinate system) and $\mubf^\text{act}$ denotes the Gaussian mean in the actor coordinate system. To transform velocities from actor to camera coordinates, $T^{\text{act}\rightarrow C}$ is the composition of the actor-to-world and world-to-camera transforms. For Gaussians that are part of the static background, $\vbf_{\text{dyn}}$ is zero. See \cref{appendix:rolling_shutter_details} for further details.

We account for the pixel velocity when culling Gaussians and checking intersections between Gaussians and tiles by increasing the approximated Gaussian extent. Instead of the square AABB used by 3DGS and \cite{seiskari2024gaussian}, we use a rectangular AABB covering $\Sigmabf^I$ at three standard deviations and add $|\vbf_I| \cdot t_\text{rs} / 2$ to its extent. Here, $t_\text{rs}$ denotes the rolling shutter duration, \ie, the time difference between the last and first row of pixels. The increase in extent corresponds to the area covered by the Gaussian mean during the rolling shutter time, assuming the sensor timestamp is in the middle of exposure. As this is not always the case, we further include a learnable time offset, describing the difference between the sensor timestamp and the middle of the exposure. Using rectangular AABBs reduces unnecessary intersections compared to \cite{kerbl20233d,seiskari2024gaussian}, especially for narrow Gaussians and Gaussians with axis-aligned velocities, the latter often being the case for side-facing cameras.

\parsection{Rasterization}
3DGS rasterizes pixels in parallel by launching a thread block per tile and assigning each thread to a pixel within the tile. For each pixel, its coordinates $\pbf=[p_u,p_v]^\text{T}$ are inferred from the block and the thread index. From this, we find the time difference between the pixel's capture time and the image's middle row as
\begin{equation}
    \label{eq:t_pix}
    t_\text{pix} = (p_v / H - 0.5) \cdot t_\text{rs}.
\end{equation}
To rasterize a pixel, we $\alpha$-blend RGB values  $\fbf_i^\text{rgb}$ and features $\fbf_i$ of the depth-sorted Gaussians intersecting the current tile
\begin{equation}
    \label{eq:alpha_blend}
    [\Fbf^\text{rgb}_\pbf, \Fbf_\pbf] = \sum_i [\fbf_i^\text{rgb},\fbf_i] \alpha_i \prod_{j=1}^{i-1}(1-\alpha_j),
\end{equation}
with $\alpha$ computed as
\begin{equation}
    \label{eq:alpha}
    \alpha_i = \sqrt{\frac{|\Sigmabf_i^I|}{|\Sigmabf_i^I+s\Ibf|}}o_i \exp\left(
    -\frac{1}{2}
    \Deltabf_i^\text{T}
    ({\Sigmabf_i^I}+s\Ibf)^{-1}
    \Deltabf_i
    \right)
\end{equation}
and
\begin{equation}
    \label{eq:delta}
    \Deltabf_i = \pbf-(\mubf^I_i + \vbf^I_i t_\text{pix}).
\end{equation}
As in \cite{seiskari2024gaussian}, we calculate the distance $\Deltabf_i$ between pixels and Gaussians using the means compensated by rolling shutter, effectively shifting Gaussians to the correct location in the image. Compared to 3DGS, in \cref{eq:alpha} we use the 3DGS+EWA~\cite{zwicker2001ewa} formulation from \cite{yu2024mip}, where $s=0.3$.

View-dependent effects are modeled using a small CNN; given the feature map $\Fbf\in\mathbb{R}^{H\times W\times D_\fbf}$, the corresponding ray directions $\dbf\in\mathbb{R}^{H\times W\times 3}$, and a camera-specific learned embedding $\ebf$, it predicts a pixel-wise affine mapping $M \in \mathbb{R}^{H\times W \times 3}$ and $\bbf \in \mathbb{R}^{H \times W \times 3}$ applied to $\Fbf^\text{rgb}$
\begin{align}
    M, \bbf &= \text{CNN}(\Fbf, \dbf, \ebf), \\
    I &= M \odot \Fbf^\text{rgb} + \bbf.
\end{align}
We found that using a small CNN instead of an MLP is essential for efficient texture modeling, \eg, for high-frequency road surfaces. In addition, inspired by \cite{rematas2022urban}, the camera-specific learned embeddings enable modeling differences in exposure.

\subsection{Lidar rendering}
\label{sec:lidar_rendering}
Lidar sensors enable self-driving vehicles to measure the distance and reflectivity (intensity) of a discrete set of points. 
They do so by emitting laser beam pulses and measuring the time of flight to determine distance and returning power for reflectivity. 
Most AD datasets deploy lidars that use several laser diodes (16-128) mounted in a vertical array, where the array of diodes is rotated to capture 360\degree~data and a single scan often takes 100 ms to capture. Therefore, we focus on this type of lidar. However, we note that modeling other types of lidars, such as solid-state lidars \cite{li2022progress}, can be easily done in our method by modifying the projection accordingly.
Our proposed lidar rendering follows the high-level steps of the image rendering, but we modify each component to accurately model lidar characteristics.

\parsection{Projection} For a posed lidar and Gaussians composed at its capture time $t$, we transform each mean and covariance from world to lidar coordinates, yielding $\mubf^L=\begin{bmatrix} x, y, z\end{bmatrix}^\text{T}$ and $\Sigmabf^L$. Raw lidar data are generally captured in spherical coordinates azimuth $\phi$, elevation $\omega$, and range $r$. Thus, we transform Gaussian means from lidar coordinates to spherical coordinates
\begin{equation}
    \label{eq:spherical_projection}
    \mubf^S
    =
    \begin{bmatrix}
    \phi \\ \omega \\ r
    \end{bmatrix}
    = 
    \begin{bmatrix}
    \arctantwo(y,x) \\ \arcsin (z/r) \\ \sqrt{x^2+y^2+z^2}
    \end{bmatrix},
\end{equation}
and convert the covariance $\Sigmabf^L$ with the Jacobian of the transform \cref{eq:spherical_projection} as $\Sigmabf^S = \Jbf^S\Sigmabf^L(\Jbf^S)^\text{T}$, where
\begin{equation}
    \Jbf^S =
    \begin{bmatrix}
        -\frac{y}{x^2 + y^2 }  & \frac{x}{x^2 + y^2} & 0 \\
        - \frac{xz}{r^2 \sqrt{x^2 + y^2}} & - \frac{yz}{r^2 \sqrt{x^2+y^2}} & \frac{\sqrt{x^2+y^2}}{ r^2} 
        \\
        \frac{x}{r} &  \frac{y}{r} &  \frac{z}{r}
    \end{bmatrix},
\end{equation}
and $\mubf^S$ and $\Sigmabf^S$ denote mean and covariance in spherical coordinates. 

Same as for the camera, we account for the rolling shutter effect by approximating its impact in sensor space
\begin{equation}
    \vbf^S = \Jbf^S (-\omegabf_L \times \mubf^L - \vbf_L + \vbf_{\text{dyn}}) \in \mathbb{R}^3.
\end{equation}
Note that $\vbf^S$ is three-dimensional as opposed to $\vbf^I$ in \cref{eq:pixel_velocity}, as we need this for accurate range rasterization. Last, we use the first two elements of $\vbf^S t_\text{rs} / 2$ to increase the extent of the rectangular AABB approximating the size of the Gaussian, and cull Gaussians outside the lidar's field-of-view.

\parsection{Tiling and sorting} At the core of 3DGS's efficiency is its tile-based rasterization. While lidars often have a fixed azimuth resolution, many use non-equidistant spacing between diodes to get higher resolution in areas of interest, resulting in nonlinear elevation resolution, as seen in \cref{fig:frontpage}. In line with this, we design tiles with a fixed number of diodes vertically and a fixed azimuth resolution horizontally. Horizontally, intersecting tiles are found by dividing azimuth coordinates by the common tile azimuth size and wrapping the results within 360\degree. Vertically, we sort the elevation tile boundaries and find intersections with a short for-loop, see \cref{appendix:lidar_rendering} for details. After tile assignment, Gaussians are sorted based on the range of their mean $\mubf^S$.

Comparing our strategy to using tiles of equal size, similar to the depth image-based rendering in \cite{chen2024omnire}, we avoid many unnecessary computations. There, the tile size must be determined from the finest resolution to ensure having at least as many threads as lidar points. Consequently, tiles in areas where the lidar is sparse would have a superficially high resolution and might even lack the corresponding lidar points to render.

\parsection{Rasterization}
Lidar data are rendered using our proposed differentiable tile-based rasterizer. Again, each tile launches one thread block, where each thread is responsible for rasterizing one lidar point. Each thread is provided the azimuth, elevation, and capture time for it to rasterize. These are either determined from lidar specifications or, for training, calculated from existing point clouds, see \cref{appendix:lidar_rendering} for details. The reason for this is that for training, one cannot expect the data to follow perfectly straight scan lines. Therefore, we cannot follow the image strategy and rely on the block and thread index to directly infer the pixel coordinates. Further, the azimuth and elevation of a lidar ray do not necessarily determine its time offset to the center of the lidar scan.

\begin{figure*}[t]
    \centering
    \includegraphics[width=\linewidth,trim={0cm 15.99cm 0cm 0cm},clip]{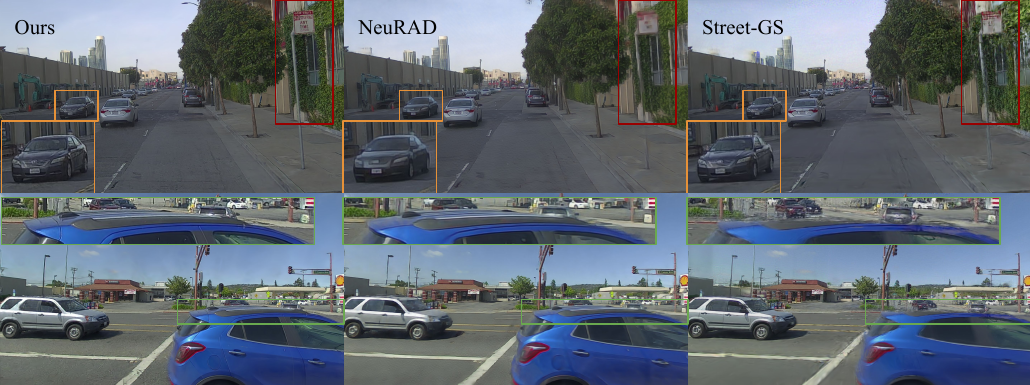}
    \caption{Compared to the baselines, {\modelname} produces sharp images with a high level of detail. Further, the bottom row highlights the superiority of our lidar rendering. Projecting lidar points into images for depth supervision, as used by previous 3DGS methods, causes line-of-sight errors and incorrect volume carving due to the pose differences between camera and lidar.}
    \label{fig:qualitative_comparison}
\end{figure*}

For each lidar point, we $\alpha$-blend the Gaussians' features $\fbf_i$ using \cref{eq:alpha_blend,eq:alpha,eq:delta}. Inspired by \cite{yu2024mip} arguing for $s$ in \cref{eq:alpha} to correspond to one pixel, we set it to correspond to the geometric mean of the lidar's vertical and horizontal beam divergence. The rendered features are concatenated with their corresponding ray direction and decoded to intensity and ray drop probability using a small MLP.

To render the expected range of a point, we $\alpha$-blend the rolling shutter corrected ranges for each Gaussian
\begin{equation}
    r_{i,\text{rs}} = r_i + \vbf_r^S t_l,
\end{equation}
where $\vbf_r^S$ is the range-component of the relative speed to the lidar and $t_l$ is the time between the capture of the current lidar point, and the middle of the lidar scan. Further, we render the median range as the rolling shutter corrected range of the first Gaussian that satisfies $\prod_{j=1}^{i}(1-\alpha_j)<0.5$ during the $\alpha$-blending in \cref{eq:alpha_blend}. While we use the expected range for training, the median range is used during inference as it, in contrast to the expected range, does not yield depths between Gaussians.

\subsection{Optimization and implementation}
\label{sec:optimization}
All model components are optimized jointly using the loss
\begin{multline}\label{eq:loss_fn}
    \mathcal{L} = \lambda_r\mathcal{L}_1 + (1-\lambda_r)\mathcal{L}_\text{SSIM} +  \lambda_\text{depth} \mathcal{L}_\text{depth} +
    \lambda_\text{los} \mathcal{L}_\text{los} + 
    \\
    \lambda_\text{intens} \mathcal{L}_\text{inten} + \lambda_\text{raydrop}\mathcal{L}_\text{BCE}  + \lambda_\text{MCMC}\mathcal{L}_\text{MCMC},
\end{multline}
where $\mathcal{L}_1$ and $\mathcal{L}_\text{SSIM}$ are L1 and SSIM losses on the rendered images. 
$\mathcal{L}_\text{depth}$ and $\mathcal{L}_\text{inten}$ are L2 losses on the rendered expected lidar range and intensity. 
$ \mathcal{L}_\text{los}$ is a line-of-sight loss, penalizing opacity before the ground truth lidar range.
$\mathcal{L}_\text{BCE}$ is a binary cross-entropy loss on the predicted ray drop probability.  
$\mathcal{L}_\text{MCMC}$ is the opacity and scale regularization used in \cite{kheradmand2024mcmc}.

We initialize the set of Gaussians using both lidar points and random points. Lidar points within bounding boxes are assigned to the corresponding actor. Further, all lidar points are projected into their closest image to set their initial color. The random points are divided into (1) uniformly sampled points within the lidar range and (2) points sampled linearly in disparity beyond the lidar range, both using random colors at initialization. Furthermore, instead of the 3DGS~\cite{kerbl20233d} splitting and densification strategy, we rely on the MCMC strategy introduced in ~\cite{kheradmand2024mcmc}. We chose MCMC partially because we find it to improve far-field rendering quality and partially because it has predictable computation requirements, as it permits choosing the maximum number of Gaussian. We use~\cite{kheradmand2024mcmc} as is, without any special treatment of Gaussians assigned to dynamic actors. See \cref{appendix:train_details} for further training details.

\parsection{Implementation}
For efficiency, we implement the forward and backward passes of the rolling shutter compensation and lidar projection and rasterization using custom CUDA kernels. {\modelname} builds upon the collaborative open-source frameworks \texttt{gsplat}~\cite{ye2024gsplat} and \texttt{neurad-studio}~\cite{neuradstudio}, and we hope that our contributions can benefit future research on multi-modal simulation, not only within the AD community. We train {\modelname} for 30,000 iterations using the Adam optimizer, which takes an hour on a single NVIDIA A100. See \cref{appendix:train_details} for further details.

%% file: sec/4_experiments.tex
\section{Experiments}
\label{sec:Experiments}
To validate the robustness of our method, we evaluate it across multiple popular AD datasets, using the same set of hyperparameters. For each dataset, we compare {\modelname} to the best performing NeRF and 3DGS-based methods tailored to AD data on novel view synthesis for both camera and lidar data. Further, we ablate key elements of {\modelname} and measure the impact they have on sensor realism and rendering speed.

\input{assets/image_nvs_table}

\input{assets/lidar_nvs_table}
\input{assets/recon_table}

\parsection{Datasets}
We perform experiments on PandaSet~\cite{pandaset}, Argoverse2~\cite{Argoverse2} and nuScenes~\cite{caesar2020nuscenes}. These datasets vary in lidars, camera appearance, and resolution. For PandaSet and nuScenes, we use the six available cameras. For Argoverse2, we use the seven ring cameras, leaving out the black-and-white stereo cameras. Further, some cameras are cropped slightly to remove views of the ego-vehicle, such as the hood and the trunk. We use the same 10 challenging sequences per dataset as in \cite{tonderski2024neurad}, containing different types of illumination, dynamic actors, and velocities. We also follow the evaluation protocol of \cite{tonderski2024neurad} \ie, using full resolution data, and, for novel view synthesis (NVS), use every other frame for training and the remaining frames for hold-out validation. We believe that this setting is more challenging than what is often encountered in previous work ~\cite{chen2024omnire,yan2024street,chen2023periodic}, where fewer datasets are considered, images are downsampled, and a larger fraction of the data is used for training.

\parsection{Baselines}
We compare our results with the best performing NeRF and 3DGS methods on AD data, namely, UniSim~\cite{unisim}, NeuRAD~\cite{tonderski2024neurad}, PVG~\cite{chen2023periodic}, Street Gaussians~\cite{yan2024street} and OmniRe~\cite{chen2024omnire}. Note that all methods are designed for dynamic scenes and use lidar data for supervision. We use NeuRAD's official implementation \texttt{neurad-studio}~\cite{neuradstudio} and their version of UniSim, as well as OmniRe's official implementation \texttt{drivestudio}~\cite{drivestudio} and their implementations of PVG and Street Gaussians. We use \texttt{drivestudio} to generate the required sky masks for Street Gaussians and OmniRe, as well as extract human pose estimates, which takes about 10 and 30 minutes per sequence on an A100. 

Although point cloud rendering is shown in~\cite{chen2024omnire}, there exists no feature for it in \texttt{drivestudio}. To render point clouds in training and validation views, we render depth images by placing 6 virtual cameras in the lidar origin, each with a horizontal FOV of 60\degree. Their vertical FOV is set to capture all lidar points in the sweep, and the focal length is the median focal length for each dataset. We project ground truth lidar points into these images, fetch each point's depth, and project them back into 3D. As \texttt{drivestudio} lacks functionality for inferring missing lidar points, these baselines are evaluated only for existing points. In contrast, {\modelname} is evaluated on the more challenging task of predicting the range also for missing rays and must filter them based on the predicted ray drop probability.
See \cref{appendix:baseline_details} for further baseline details.
\begin{figure}[t]
    \centering
    \includegraphics[width=\linewidth,trim={0cm 18.01cm 0cm 0cm},clip]{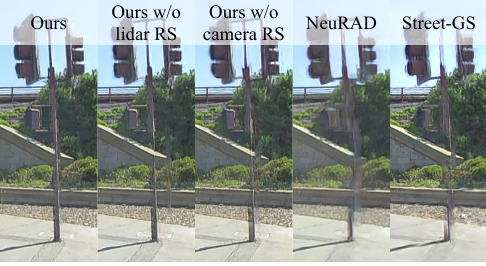}
    \caption{Removing our rolling shutter modeling compensation leads to inaccurate geometries and inconsistencies in the learning.}
    \label{fig:qualitative_rs}
\end{figure}

\subsection{Image rendering}
\parsection{Novel view synthesis}
We assess the rendering quality of images using the standard NVS metrics PSNR, SSIM \cite{wang2004image}, and LPIPS \cite{zhang2018unreasonable} on hold-out validation images in \cref{tab:results_table_nvs_image}. We measure speed using resolution-agnostic megapixels per second. {\modelname} consistently outperforms existing 3DGS methods in terms of image quality with a large margin for the three considered datasets. Further, our method improves upon NeuRAD's SOTA results in all NVS metrics, while rendering images an order of magnitude faster. We provide a qualitative comparison in \cref{fig:qualitative_comparison}.

\parsection{Reconstruction}
To measure the upper limit of the modeling capacity of the methods, we report reconstruction metrics on PandaSet in \cref{tab:results_table_reconstruction}. Here, methods train on all data in a sequence and evaluate on the same views. Further, we enable sensor pose optimization for all methods to account for any pose inaccuracies. Comparing to the NVS setting, all methods except UniSim show improvements with the additional training data. {\modelname} achieves SOTA results while rendering $\times10$ faster than the previous best method. Further, we note that {\modelname}'s performance on hold-out validation images in \cref{tab:results_table_nvs_image} is on par with previous 3DGS methods' results for reconstruction in \cref{tab:results_table_reconstruction}.

\parsection{Extrapolation}
We investigate {\modelname}'s ability to generalize to views that differ significantly from what is encountered during training. We use the models trained on every other frame and follow the three settings in \cite{tonderski2024neurad}: shift the ego-vehicle horizontally, shift the ego-vehicle vertically, and apply shifts and rotation to all dynamic actors. We report the Fréchet distance using DINOv2~\cite{oquab2024dinov} features, as these have been shown~\cite{stein2024exposing} to align better with human perception than using Inception-v3~\cite{szegedy2016rethinking}. However, we note that using Inception-v3 features instead does not change the model ranking or our conclusions. \cref{tab:fid_table} shows {\modelname}'s ability to learn meaningful representations for generalization, clearly outperforming other 3DGS methods.

\input{assets/fid_table}

\subsection{Lidar rendering}
We measure the quality of our lidar point clouds using the same metrics as in \cite{tonderski2024neurad}, \ie, median squared depth error, RMSE intensity error, ray drop accuracy, and chamfer distance, see \cref{tab:results_table_nvs_lidar} for NVS and \cref{tab:results_table_reconstruction} for reconstruction. Note that we do not report intensity or ray drop accuracy for the 3DGS baselines, as they have no way of inferring these. Further, it was not possible to produce Argoverse2 point cloud results for these baselines due to the memory requirements of \texttt{drivestudio} not scaling well with the increase in number of images and higher image resolution. We find {\modelname} to be on par with the accurate ray-tracing formulation of NeuRAD in terms of point cloud quality, while running up to $\times18$ faster. Furthermore, our lidar rendering approach is superior to the naive depth image-based method used for 3DGS baselines, both in terms of speed and quality. 

\input{assets/ablations_average}

\begin{figure}[t]
    \centering
    \includegraphics[width=\linewidth,trim={0cm 20.22cm 0cm 0cm},clip]{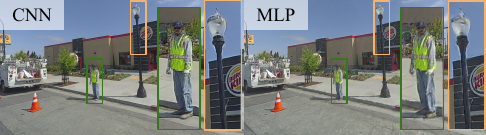}
    \caption{The CNN decoder improves sharpness and is more true to color than the MLP decoder.}
    \vspace{-6mm}
    \label{fig:qualitative_cnn_mlp}
\end{figure}

\subsection{Ablations}
We validate the effectiveness of key components of our method by measuring their impact on NVS metrics in \cref{tab:average_ablations}. Metrics are averaged over all three datasets to avoid any dataset-specific biases. We see that removing the rolling shutter modeling of the camera (a) or lidar (f) reduces NVS performance, and that leaving out optimization of sensor velocities (e) negatively impacts our depth predictions. While the impact of rolling shutter modeling on quantitative metrics is somewhat modest, we qualitatively find that it is important for capturing thin structures and edges, see \cref{fig:qualitative_rs}. Removing these components leads to line-of-sight errors, which erroneously can cut through objects. Moreover, we note the efficiency of our rolling shutter algorithms, as removing them barely improves rendering speeds. 

We also find that using a CNN instead of an MLP for image decoding (c) clearly improves image realism, without any major runtime impact, see \cref{fig:qualitative_cnn_mlp} for an example. Further, we verify that using median depth instead of expected depth during inference is important for realistic lidar data in terms of Chamfer distance (h). Last, we note that MCMC~\cite{kheradmand2024mcmc} and EWA antialiasing~\cite{yu2024mip,zwicker2001ewa} both improve our performance, with the antialiasing having the largest impact on perceptual quality in terms of LPIPS.

%% file: assets/image_nvs_table.tex
\begin{table}[t]
    \centering
    \caption{NVS results for image, over three datasets. \colorbox{tabfirst}{First}, \colorbox{tabsecond}{second}, \colorbox{tabthird}{third}.}
    \resizebox{0.85\linewidth}{!}{
    \begin{tabular}{ll ccccc}
    &     &  PSNR $\uparrow$ & SSIM $\uparrow$ & LPIPS $\downarrow$ & MP/s $\uparrow$ & Train (H) $\downarrow$\\ \midrule
    \multirow{6}{*}{\rotatebox[origin=c]{90}{\shortstack[c]{\small PandaSet}}}
& UniSim              & 23.12                     & 0.682                      & 0.360                        &                       11.5  & \cellcolor{tabfirst}0.9\\
& NeuRAD              & \cellcolor{tabsecond}25.80& \cellcolor{tabsecond}0.753 & \cellcolor{tabsecond}0.250   &                        9.7  & \cellcolor{tabthird}1.3\\
& PVG                 & 24.01                     & 0.712                      & 0.452                        & \cellcolor{tabsecond}28.2   & 1.6\\ 
& Street-GS           & \cellcolor{tabthird}24.73 & 0.745\cellcolor{tabthird}  & \cellcolor{tabthird}0.314    & \cellcolor{tabthird}22.8    & 2.3\\
& OmniRE              & \cellcolor{tabthird}24.71 & 0.745\cellcolor{tabthird}  & \cellcolor{tabthird}0.315    &                      16.5   & 2.6\\
& {\modelname} (ours) & \cellcolor{tabfirst}26.76 & \cellcolor{tabfirst}0.815  & \cellcolor{tabfirst}0.193    & \cellcolor{tabfirst}121.5   & \cellcolor{tabsecond}1.0\\
    \midrule                                                                                                                                        
    \multirow{6}{*}{\rotatebox[origin=c]{90}{\shortstack[c]{\small Argo2}}}                                                           
& UniSim              &                    22.35   &                    0.655   &                    0.458    &                      10.0   & \cellcolor{tabfirst}1.1\\ 
& NeuRAD              & \cellcolor{tabsecond}26.18 & \cellcolor{tabthird}0.721  & \cellcolor{tabsecond}0.310  &                      9.7    & \cellcolor{tabthird}1.3\\ 
& PVG                 &                      24.47 &                      0.712 &                    	0.502 & \cellcolor{tabsecond}48.0   & 2.5\\ 
& Street-GS           &                      25.52 & \cellcolor{tabsecond}0.754 & \cellcolor{tabthird}0.374   &  \cellcolor{tabthird}38.7   & 2.6\\ 
& OmniRE              & \cellcolor{tabthird}25.61  & \cellcolor{tabsecond}0.753 & \cellcolor{tabthird}0.375   &                      23.9   & 3.0\\ 
& {\modelname} (ours) & \cellcolor{tabfirst}28.42  & \cellcolor{tabfirst}0.826  & \cellcolor{tabfirst}0.270   & \cellcolor{tabfirst}134.5   & \cellcolor{tabsecond}1.2\\ 
    \midrule                                                                                                                                           
    \multirow{6}{*}{\rotatebox[origin=c]{90}{\shortstack[c]{\small nuScenes}}}                                                              
& UniSim              &                     22.66  &                     0.743  &                     0.414   &                      13.3  	& \cellcolor{tabfirst}0.7\\ 
& NeuRAD              & \cellcolor{tabsecond}26.17 & \cellcolor{tabsecond}0.790 & \cellcolor{tabsecond}0.312  &                      11.8  	& \cellcolor{tabsecond}1.1\\ 
& PVG                 &                     24.94  & \cellcolor{tabthird}0.768  &                     0.497   & \cellcolor{tabthird}18.8    & 1.4\\ 
& Street-GS           & \cellcolor{tabthird}25.54  & \cellcolor{tabsecond}0.799 & \cellcolor{tabthird}0.375   & \cellcolor{tabsecond}21.5   & 1.6\\ 
& OmniRE              &                     25.50  & \cellcolor{tabsecond}0.798 &                     0.386   &                      16.3   & 1.9\\ 
& {\modelname} (ours) & \cellcolor{tabfirst}27.54  & \cellcolor{tabfirst}0.849  &  \cellcolor{tabfirst}0.302  & \cellcolor{tabfirst}106.1   & \cellcolor{tabthird}1.2\\ 
    \bottomrule                                                                                                                                                                                                                                                                                                                      
    \end{tabular}                                                                                                                                                                                                                                                                                                                      
    }                                                                                                                                                                                                                                                                                                                      
    \label{tab:results_table_nvs_image}                                                                                                                                                                                                                                                                                                                      
\end{table}

%% file: assets/lidar_nvs_table.tex
\begin{table}[t]
    \centering
    \caption{NVS results for lidar, over three datasets. $^\mathsection$without missing points. \colorbox{tabfirst}{First}, \colorbox{tabsecond}{second}, \colorbox{tabthird}{third}.}
    \resizebox{0.85\linewidth}{!}{
    \begin{tabular}{ll ccccc}
    &     &  Depth $\downarrow$ & Intensity $\downarrow$ & Drop acc. $\uparrow$ & CD $\downarrow$ & MR/s $\uparrow$\\ \midrule
    \multirow{6}{*}{\rotatebox[origin=c]{90}{\shortstack[c]{\small PandaSet}}}
& UniSim              & \cellcolor{tabsecond}0.08 & \cellcolor{tabthird}0.086  &    -                      & \cellcolor{tabthird}10.3$^\mathsection$     & \cellcolor{tabsecond}0.9   \\
& NeuRAD              & \cellcolor{tabfirst}0.01  & \cellcolor{tabsecond}0.063 & \cellcolor{tabsecond}96.2 & \cellcolor{tabsecond}1.9                    & \cellcolor{tabsecond}1.1   \\
& PVG                 & 38.74                     &   -                        &    -                      & 125.2$^\mathsection$                        & \cellcolor{tabthird}0.2    \\
& Street-GS           & 6.18                      &   -                        &    -                      & 37.3$^\mathsection$                         & \cellcolor{tabthird}0.2    \\
& OmniRE              & \cellcolor{tabthird}2.88  &   -                        &    -                      & 29.8$^\mathsection$                         & \cellcolor{tabthird}0.2    \\
& {\modelname} (ours) & \cellcolor{tabfirst}0.01  & \cellcolor{tabfirst}0.059  & \cellcolor{tabfirst}96.7  & \cellcolor{tabfirst}1.6                     & \cellcolor{tabfirst}19.5   \\
    \midrule                                                                                                                                        
    \multirow{3}{*}{\rotatebox[origin=c]{90}{\shortstack[c]{\small Argo2}}}                                                           
& UniSim              & \cellcolor{tabsecond}0.18 & \cellcolor{tabthird}0.081  &                    -	  & \cellcolor{tabthird}29.2$^\mathsection$    & \cellcolor{tabsecond}0.7   \\
& NeuRAD              & \cellcolor{tabfirst}0.02  & \cellcolor{tabsecond}0.058 & \cellcolor{tabsecond}92.2 & \cellcolor{tabfirst}2.6                    & \cellcolor{tabsecond}0.9  \\ 
& {\modelname} (ours) & \cellcolor{tabfirst}0.02  & \cellcolor{tabfirst}0.052  & \cellcolor{tabfirst}92.6  & \cellcolor{tabsecond}2.8                   & \cellcolor{tabfirst}9.5   \\ 
    \midrule                                                                                                                                           
    \multirow{6}{*}{\rotatebox[origin=c]{90}{\shortstack[c]{\small nuScenes}}}                                                              
& UniSim              & \cellcolor{tabthird}0.06   & \cellcolor{tabthird}0.063  &                       -   &                     52.2$^\mathsection$   &\cellcolor{tabsecond}1.0    \\
& NeuRAD              & \cellcolor{tabfirst}0.01   & \cellcolor{tabsecond}0.042 & \cellcolor{tabsecond}93.1 & \cellcolor{tabthird}6.3	               &\cellcolor{tabsecond}1.1     \\
& PVG                 &                     10.34  &                       -    &                       -   &                     77.9$^\mathsection$   & \cellcolor{tabthird}0.02   \\
& Street-GS           &                     1.56   &                       -    &                       -   & \cellcolor{tabsecond}4.5$^\mathsection$   & \cellcolor{tabthird}0.04   \\
& OmniRE              &                     1.71   &                       -    &                       -   & \cellcolor{tabsecond}4.6$^\mathsection$ 	& \cellcolor{tabthird}0.03  \\
& {\modelname} (ours) & \cellcolor{tabsecond}0.02  & \cellcolor{tabfirst}0.036  & \cellcolor{tabfirst}93.8  & \cellcolor{tabfirst}1.7	               & \cellcolor{tabfirst}5.7    \\
    \bottomrule                                                                                                                                                                                                                                                                                                                      
    \end{tabular}                                                                                                                                                                                                                                                                                                                      
    }                                                                                                                                                                                                                                                                                                                      
    \label{tab:results_table_nvs_lidar}                                                                                                                                                                                                                                                                                                                      
\end{table}

%% file: assets/recon_table.tex
\begin{table*}[t]
    \centering
    \caption{Reconstruction results for image and lidar point clouds on PandaSet. $^\mathsection$without missing points. \colorbox{tabfirst}{First}, \colorbox{tabsecond}{second}, \colorbox{tabthird}{third}.}
    \resizebox{0.85\linewidth}{!}{
    \begin{tabular}{ll ccc cccc ccc}
    
    \multicolumn{2}{c}{} & \multicolumn{3}{c}{Image} & \multicolumn{4}{c}{Lidar} & \multicolumn{3}{c}{Efficiency} \\ \cmidrule(lr){3-5} \cmidrule(lr){6-9} \cmidrule(lr){10-12}
    &     &  PSNR $\uparrow$ & SSIM $\uparrow$ & LPIPS $\downarrow$ 
    & Depth $\downarrow$ & Intensity $\downarrow$ & Drop acc. $\uparrow$ & CD $\downarrow$ 
    & Camera MP/s $\uparrow$ & Lidar MR/s $\uparrow$ & Train time (H) $\downarrow$ \\ \midrule

    \multirow{6}{*}{\rotatebox[origin=c]{90}{\shortstack[c]{\small PandaSet}}}
    & UniSim                &                    23.18   &                    0.684   &                    0.362   & \cellcolor{tabthird}0.06  & \cellcolor{tabthird}0.087  &                      -    & \cellcolor{tabthird}10.2$^\mathsection$  &                    11.1    & \cellcolor{tabthird}0.8  & \cellcolor{tabfirst}0.8\\
    & NeuRAD                & \cellcolor{tabsecond}26.76 & \cellcolor{tabthird}0.778  & \cellcolor{tabsecond}0.233 & \cellcolor{tabsecond}0.03   & \cellcolor{tabsecond}0.074 & \cellcolor{tabsecond}96.1 & \cellcolor{tabsecond}1.7                  &                    9.8     & \cellcolor{tabsecond}1.1 & \cellcolor{tabthird}1.3\\
    & PVG                   &                    25.35   &                    0.752   &                    0.423   &                    53.18   &                      -     &                      -    &                    150.8$^\mathsection$  & \cellcolor{tabsecond}73.8  &                    0.2   & 1.6\\ 
    & Street-GS             & \cellcolor{tabthird}26.68  & \cellcolor{tabsecond}0.808 & \cellcolor{tabthird}0.293  &                      3.27  &                      -     &                      -    &                    27.9$^\mathsection$   & \cellcolor{tabthird}23.9   &                    0.2   & 2.2\\
    & OmniRE                & \cellcolor{tabthird}26.68  & \cellcolor{tabsecond}0.808 & \cellcolor{tabthird}0.295  &                    3.80    &                      -     &                      -    &                    33.5$^\mathsection$   &                    16.6    &                    0.2   & 2.4\\
    & {\modelname} (ours)   & \cellcolor{tabfirst}29.74  & \cellcolor{tabfirst}0.893  & \cellcolor{tabfirst}0.175  & \cellcolor{tabfirst}0.02   & \cellcolor{tabfirst}0.053  & \cellcolor{tabfirst}97.4  & \cellcolor{tabfirst}1.6                 & \cellcolor{tabfirst}114.4  & \cellcolor{tabfirst}17.4 & \cellcolor{tabsecond}1.0\\
    \bottomrule
    \end{tabular}
    }
    \label{tab:results_table_reconstruction}
\end{table*}

%% file: assets/fid_table.tex
\begin{table}[t]
    \centering
    \caption{FD$_\text{DINOv2}$ scores when shifting pose of ego vehicle or actors. \colorbox{tabfirst}{First}, \colorbox{tabsecond}{second}, \colorbox{tabthird}{third}.}
    \resizebox{0.85\linewidth}{!}{
    \begin{tabular}{l c c c c c c c}
    \toprule
    &     & \multicolumn{3}{c}{Ego lane shift} & Ego vert. shift & \multicolumn{2}{c}{Actor shift} \\
    &     & 0m & 2m & 3m & 1m & Rot. & Trans. \\
    \midrule
    \multirow{5}{*}{\rotatebox[origin=c]{90}{\shortstack[c]{\small PandaSet}}}
& NeuRAD              &\cellcolor{tabsecond}377.4 &\cellcolor{tabsecond}540.8 &\cellcolor{tabfirst}629.2  &\cellcolor{tabfirst}571.0  &\cellcolor{tabsecond}424.8 &\cellcolor{tabsecond}420.2 \\
& PVG                 &                    618.8  &                    782.7  &                    885.6  &                    913.9  &                    618.8  &                    618.8 \\ 
& Street-GS           &\cellcolor{tabthird}497.1  &\cellcolor{tabthird}718.6  &\cellcolor{tabthird}824.2  &\cellcolor{tabthird}868.7  &\cellcolor{tabthird}554.7  &\cellcolor{tabthird}558.0 \\
& OmniRE              &                    504.5  &                    723.2  &                    833.5  &                    872.8  &                    569.6  &                    573.2 \\
& {\modelname} (ours) &\cellcolor{tabfirst}276.3  &\cellcolor{tabfirst}520.7  &\cellcolor{tabsecond}647.2 &\cellcolor{tabsecond}613.2 &\cellcolor{tabfirst}355.1  &\cellcolor{tabfirst}345.1 \\
    \bottomrule
    \end{tabular}
    }
    \label{tab:fid_table}
\end{table}

%% file: assets/ablations_average.tex
\begin{table}[t]
    \centering
    \caption{NVS results averaged over 10 sequences from PandaSet, nuScenes, and Argoverse2 when \textit{removing} model components.}
    \addtolength{\tabcolsep}{-2pt}
    \vspace{-2mm}
    \resizebox{0.99\linewidth}{!}{
    \begin{tabular}{clccccccc}
    \toprule
        &                   & \multirow{2}{*}{PSNR $\uparrow$}   & \multirow{2}{*}{SSIM $\uparrow$} & \multirow{2}{*}{LPIPS $\downarrow$} & \multirow{2}{*}{Depth $\downarrow$} & \multirow{2}{*}{CD $\downarrow$} & \multirow{2}{*}{\shortstack[c]{Camera\\MP/s $\uparrow$}} & \multirow{2}{*}{\shortstack[c]{Lidar\\MR/s $\uparrow$}}  \\
        \\
       & Full model        & 27.58 & 0.830 & 0.254 & 0.02 & 2.4 & 119.6 & 11.6  \\ \midrule
    a) & Camera RS           & 27.46 & 0.827 & 0.258 & 0.03 & 2.3 & 121.4 & 11.4   \\
    b) & CNN decoder         & 27.09 & 0.822 & 0.266 & 0.04 & 2.4 & 122.4 & 11.6   \\
    c) & EWA antialias                 & 27.46 & 0.824 & 0.281 & 0.06 & 1.7 & 120.5 & 11.7   \\
    d) & Appearance emb.     & 27.30 & 0.823 & 0.264 & 0.03 & 2.6 & 122.7 & 11.9   \\
    e) & Velocity opt. & 27.58 & 0.829 & 0.257 & 0.04 & 2.4 & 118.9 & 11.4   \\
    f) & Lidar RS            & 27.39 & 0.827 & 0.258 & 0.05 & 2.4 & 119.1 & 12.2   \\
    g) & MCMC                & 27.30 & 0.824 & 0.260 & 0.02 & 2.0 & 128.7 & 14.9   \\
    h) & Median depth        & 27.58 & 0.830 & 0.254 & 0.02 & 4.9 & 119.6 & 11.6  \\
    \bottomrule
    \end{tabular}
    }
    \addtolength{\tabcolsep}{2pt}
    \label{tab:average_ablations}
    \vspace{-2mm}
\end{table}

%% file: sec/5_conclusion.tex
\section{Conclusion}
\label{sec:conclusion}
In this work, we have proposed {\modelname}, the first method for rendering camera and lidar data from 3D Gaussians. We combined accurate sensor modeling with efficient algorithms to achieve state-of-the-art novel view synthesis results while improving run-time by more than an order of magnitude. We hope that {\modelname} will advance the scalability and fidelity of simulation environments, facilitating safer and more cost-effective testing of AD systems.

\parsection{Limitations and future work}
{\modelname} is currently limited to modeling all dynamic actors as rigid. Drawing inspiration from recent advances in human reconstruction~\cite{10.1007/978-3-031-73229-4_26,kocabas2024hugs,moreau2024human} can provide inspiration how to overcome this limitation in future research.
In addition, we believe that incorporating data-driven priors such as diffusion models~\cite{wu2024reconfusion,gao2024catd,esser2024scaling} is a promising direction to improve the extrapolation performance, and for changing appearance attributes.

\subsection*{Acknowledgements}
We thank Adam Lilja, William Ljungbergh, and Adam Tonderski for valueable feedback.
This work was partially supported by the Wallenberg AI, Autonomous Systems and Software Program (WASP) funded by the Knut and Alice Wallenberg Foundation. Computational resources were provided by NAISS at \href{https://www.nsc.liu.se/}{NSC Berzelius}, partially funded by the Swedish Research Council, grant agreement no. 2022-06725.

%% file: sec/6_suppl.tex
\clearpage
\setcounter{page}{1}
\maketitlesupplementary
\appendix
In the supplementary material, we provide implementation details for our method and baselines, evaluation details, and additional results.
In \cref{appendix:lidar_rendering}, we describe the details of the lidar rendering.
In \cref{appendix:train_details}, we describe our training setup more closely and provide hyperparameters.
In \cref{appendix:baseline_details}, we describe the process of how we generate point clouds with our baselines.
In \cref{appendix:rolling_shutter_details}, we provide additional details and explanations of our rolling shutter modeling. 
In \cref{appendix:results}, we give additional qualitative examples from our method and baselines. 
Last, in \cref{appendix:eval_details} we provide evaluation details.

\section{Lidar rendering details}
\label{appendix:lidar_rendering}
\subsection{Lidar tiling}
The tiling for lidar rasterization is done such that each tile has the same number of lidar points to rasterize.
Horizontally, we define each tile to cover $N_\phi$ lidar points, corresponding to $N_\phi \cdot \text{res}_\phi$ degrees, where $\text{res}_\phi$ is the azimuth resolution.
Vertically, each tile covers $N_\omega$ elevation channels, where the elevation channels are defined in lidar specifications.
We set the elevation tile boundaries to lie between the elevation channels.
Thus, each lidar tile can rasterize $N_\phi \cdot N_\omega$ lidar points. 
We set $N_\phi=32$ and $N_\omega=8$ so that $N_\phi \cdot N_\omega=256$, analogous to the $16\times16=256$ pixels in the image tiles. 
Further, a lidar point cloud corresponds to $M_\phi = \ceil{360\degree / (N_\phi \cdot \text{res}_\phi)}$ tiles horizontally, 
and $M_\omega = N_\text{beams} / N_\omega$ vertically, where $N_\text{beams}$ is the number of lidar beams/channels.

As described in \cref{sec:lidar_rendering}, to find intersections between all Gaussians and tiles, each Gaussian is defined by an AABB $[(\phi_\text{low}, \omega_\text{low}), (\phi_\text{high},  \omega_\text{high})]$ centered around its 2D mean in spherical coordinates $\mubf^S$.
The extent of the AABB corresponds to the projected $3-\sigma$ size of the Gaussian, which further is expanded by acknowledging its velocity.
For each Gaussian, we convert these spherical coordinates to tile coordinates, \eg, $(2,3)$ denotes the tile that is second horizontally and third vertically. 
This yields an AABB in tile coordinates $[(\Phi_\text{low}, \Omega_\text{low}), (\Phi_\text{high},  \Omega_\text{high})]$. 
By computing the lower limits inclusively, and the upper limits exclusively, the area of the AABB expressed in tile coordinates corresponds to the number of intersections for that Gaussian.

When finding the azimuth tile coordinates, we must account for the wrapping of angles. 
Although the 2D means of Gaussians are bound to $[0\degree, 360\degree)$, their AABB coordinates are not. 
In addition, the last azimuth tile might extend beyond $360\degree$ as $\phi_\text{max}=M_\phi\cdot N_\phi \cdot \text{res}_\phi \geq 360\degree$.
When wrapping the angles, this can create some overlap between the first and last column of tiles.
Since this overlapping area has no corresponding lidar points for the last column of tiles (they are already taken care of by the first column of tiles), it should not be considered for intersections.
Thus, the lower limit is found as
\begin{equation}
    \Phi_\text{low} = 
    \begin{cases}
        \floor{\frac{\phi_\text{low}}{ N_\phi \cdot \text{res}_\phi}} &\text{ if } \phi_\text{low} \geq 0,
        \\
        \floor{\frac{(\phi_\text{low} + 360) - \phi_\text{max}}{ N_\phi \cdot \text{res}_\phi}} &\text{otherwise,}
    \end{cases}
\end{equation}
and upper limit as
\begin{equation}
    \Phi_\text{high} = 
    \begin{cases}
        \ceil{\frac{\phi_\text{high}}{ N_\phi \cdot \text{res}_\phi}} &\text{ if } \phi_\text{high} \leq 360,
        \\
        \ceil{\frac{\phi_\text{high} \Mod{360}}{ N_\phi \cdot \text{res}_\phi}} + M_\phi &\text{otherwise.}
    \end{cases}
\end{equation}
Given $\Phi_\text{low}$ and $\Phi_\text{high}$, the number of tiles covered horizontally is $\Phi_\text{high} - \Phi_\text{low}$. As for the image-case, Gaussians are duplicated for each intersection and associated with a unique identifier based on the intersecting tile and the depth of the Gaussian. This unique identifier is used for global sorting. However, before creating this identifier, the azimuth tile coordinates of intersections are wrapped to $[0, M_\phi)$
\begin{align}
    \Phi_\text{low, wrapped} &= (\Phi_\text{low} + M_\phi) \Mod{M_\phi},
    \\
    \Phi_\text{high, wrapped} &= (\Phi_\text{high} + M_\phi) \Mod{M_\phi}.
\end{align}

For the elevation tile coordinates, we loop over the sorted elevation boundaries and set $\Omega_\text{low}$ to the last boundary that is smaller than $\omega_\text{low}$ and $\Omega_\text{high}$ to the last boundary that is larger than $\omega_\text{high}$.

\subsection{Lidar points to rasterization points}
During training and evaluation, we provide the azimuth and elevation values used for rasterization based on the collected lidar data.
We begin by removing any ego-motion compensation by expressing the lidar points' coordinates relative to the sensor pose at the time of capture. 
For this, we assume a linear motion during the lidar scan's capture.
Next, for each lidar point $\xbf=[x,y,z]^\text{T}$, we convert it to spherical coordinates
\begin{equation}
    \xbf^S
    =
    \begin{bmatrix}
    \phi \\ \omega \\ r
    \end{bmatrix}
    = 
    \begin{bmatrix}
    \arctantwo(y,x) \\ \arcsin (z/r) \\ \sqrt{x^2+y^2+z^2}
    \end{bmatrix}.
\end{equation}
Each point is then mapped to a single tile, using the same approach as for the Gaussians, while assuming that the extent of each lidar point is zero.

In some cases, this process can assign slightly more than 256 lidar points to a tile. 
This occurs, for instance, if the linear motion assumption for the ego-motion removal is violated.
During training, we shuffle points for each iteration and discard any points beyond 256.
For evaluation, we run multiple rasterization passes and concatenate the results.

For tiles with less than 256 points, we still spawn 256 threads for rasterization. Similar to how threads that have reached sufficient accumulation only help loading new batches of Gaussians into shared memory, we use threads without assigned lidar points for the same purpose.

\section{Training details}
\label{appendix:train_details}
In this section, we present details of our model and how we train it.

\parsection{Optimization}
All parameters of our model are optimized jointly for 30,000 steps, using the Adam \cite{adam} optimizer. Learning rates for the different parameters are reported in \cref{tab:lr_table}, and are scheduled using an exponential decay scheduler when applicable. Following \cite{kerbl20233d}, we start the optimization with images 4 times smaller than the original resolution and upsample with a factor of 2 after 3,000 and 6,000 steps.

\parsection{Initialization}
Gaussians are initialized from a mix of lidar points and random points. We use a maximum of 2M lidar points for the static world and add 500 points, drawn randomly within the box, for each actor. In addition to the lidar points, we also initialize 60,000 Gaussians from random points. Half of these points are sampled uniformly within the lidar range. The other half is created from uniformly sampled directions and distances sampled linearly in inverse distance beyond the lidar range, up to a maximum distance of 10 kilometers. Gaussians created from lidar points are initialized with the color retrieved from projecting the point into the closest image, while Gaussians created from random points are initialized with random colors. The scale of a Gaussian is initialized as 20$\%$ of the average distance to its three nearest neighbors, and the opacity is initialized to $0.5$. 

\parsection{Loss hyperparameters}
Our model is optimized by minimizing \cref{eq:loss_fn} with $\lambda_r = 0.8$, $\lambda_{\text{depth}} = 0.1$, $\lambda_{\text{los}} = 0.1$, $\lambda_{\text{intens}} = 1.0$, and $\lambda_{\text{raydrop}} = 0.1$. Except for $\lambda_r$, for which we use the same value as in ~\cite{kerbl20233d}, all hyperparameters are set heuristically. The MCMC loss in \cref{eq:loss_fn} is adapted from \cite{kheradmand2024mcmc} and consists of an opacity regularization term and a scale regularization term
\begin{equation}
    \lambda_{\text{MCMC}} \mathcal{L}_{\text{MCMC}} = \lambda_o \sum_i  \left| o_i \right| + \lambda_\Sigma  \sum_{ij} \left| \sqrt{ \text{eig}_j (\Sigma_i) } \right|,
\end{equation}
where we use $\lambda_{\text{o}} = 0.005$ and $\lambda_{\Sigma} = 0.001$. The line-of-sight loss for Gaussians intersecting a lidar point $p$ is implemented as
\begin{equation}
    \mathcal{L}_{\text{los}, p} = \sum_{r_i < r_p - \epsilon} \alpha_i,
\end{equation}
where $r_i$ is the range of Gaussian $i$, $r_p$ is the range of the lidar point $p$ and $\epsilon = 0.8$.

\parsection{Densification strategy}
We use the MCMC strategy introduced in \cite{kheradmand2024mcmc} with the same hyperparameters and the maximum number of Gaussians set to 5M.

\parsection{Features}
In addition to the three color channels, our Gaussians have associated features of dimension 13. We give the sensor-specific embeddings a size of 8.  The small CNN used for decoding view-dependent effects consists of two residual blocks with a hidden dimension of 32 and kernel size 3, before a final linear layer. The MLP used for decoding lidar intensity and ray drop probability is also lightweight, consisting of only 2 layers and a hidden dimension of 32.

\begin{table}
    \centering
    \caption{Learning rates (LR) for the different parameter groups. Learning rate scheduling is done using exponential decay.}
    \resizebox{0.99\linewidth}{!}{
    \begin{tabular}{l c c l}
    Parameters & Initial LR & Final LR & Warm-up steps \\
    \hline
Means & 1.6e-6 & 1.6e-6 & 0 \\
Features & 2.5e-3 & 2.5e-3 & 0 \\
Opacities & 5.0e-2 & 5.0e-2 & 0 \\
Scales & 5.0e-3 & 5.0e-3 & 0 \\
Quaternions & 1.0e-3 & 1.0e-3 & 0 \\
Sensor vel. linear  & 1.0e-3 & 1.0e-6 & 1000 \\
Sensor vel. angular  & 2.0e-4 & 1.0e-7 & 1000 \\
Cam. time to center  & 2.0e-4 & 1.0e-7 & 10000 \\
Actor trajectories  & 1.0e-3 & 1.0e-4 & 2500 \\
Sensor embeddings  & 1.0e-3 & 1.0e-3 & 500 \\
    \end{tabular}
    }
    \label{tab:lr_table}
\end{table}

\section{Baseline details}
\label{appendix:baseline_details}
The three considered 3DGS-based baselines, PVG~\cite{chen2023periodic}, Street Gaussians~\cite{yan2024street}, and OmniRe~\cite{chen2024omnire} are all implemented in the open-source repository \texttt{drivestudio}~\cite{drivestudio}. 
We modify the codebase to enable point cloud rendering by, as described in~\cite{chen2024omnire}, projecting lidar points into depth images, fetching their depth, and projecting them into 3D.

To generate a point cloud, we place 6 virtual cameras, each with a horizontal FOV of $60\degree$, in the lidar origin. The cameras are rotated such that they collectively cover $360\degree$. The focal length is set to the median horizontal focal length of each dataset. This ensures that depth images are rendered at a similar resolution as the models were trained on.

Next, the six depth images are rendered. Here, depth refers to the $\alpha$-blended $z$ coordinate of Gaussians in camera coordinates. For each lidar point, we project it into the depth image and bilinearly interpolate the nearby pixel values. The $z$-depths are then converted to ranges $t$ by dividing their values by the cosine of the angle between the direction of the lidar point and the direction of the corresponding cameras $z$-axis. The points are placed in 3D using the true lidar points origins $\obf$ and direction $\dbf$ as $\obf + \dbf t$.

\section{Rolling shutter details}
\label{appendix:rolling_shutter_details}
\begin{figure}[t]
    \centering
    \includegraphics[width=0.9\linewidth, trim={0cm 13.0cm 0cm 0cm},clip]{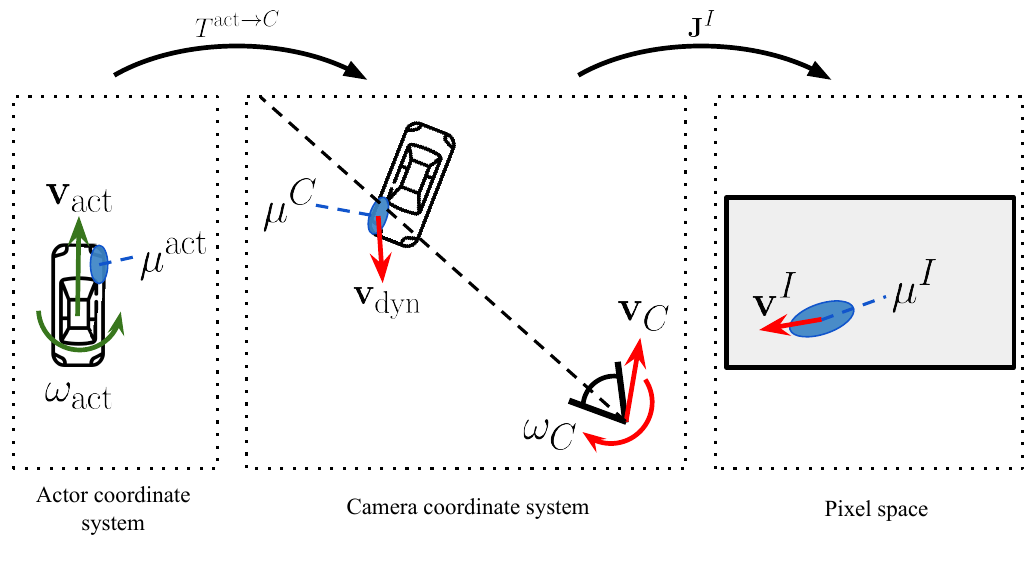}
    \caption{Visualization of an example of the components in the pixel velocity equation.}
    \label{fig:pixel_velocity_visualization}
\end{figure}

Our rolling shutter compensation is computed from approximated velocities in image space, referred to as pixel velocities. We assume the movement of a sensor $C$ at time $t$ to be modeled by a linear velocity $\vbf_C$ and an angular velocity $\omegabf_C$, expressed in the sensor's coordinate system, as exemplified in \cref{fig:pixel_velocity_visualization}. For a static Gaussian $i$, expressed in the coordinate system of this sensor, we can thus describe its velocity relative to the sensor as
\begin{equation} \label{eq:vel_rel_cam}
    \vbf_{i, \text{static}}^C = -(\omegabf_C \times \mubf_i^C + \vbf_C),
\end{equation}
where $\mubf_i^C$ is the mean of the Gaussian, expressed in the coordinate system of sensor $C$. Note that the sign of the velocity is negative because we are interested in the velocity of the Gaussian \textit{relative} to the sensor.

If the Gaussian is associated with a dynamic actor, we must also consider the velocity contribution induced by the actor's movement. To this end, we introduce the ``actor coordinate system", which is aligned with the actor's current pose but at a fixed location and rotation in the world coordinate system. For an actor modeled by a linear velocity $\vbf_\text{act}$ and an angular velocity $\omegabf_\text{act}$, as exemplified in \cref{fig:pixel_velocity_visualization}, we can describe the velocity of Gaussian $i$ as
\begin{equation}
    \vbf_{i, \text{dyn}}^{\text{act}} = \omegabf_\text{act} \times \mubf_i^\text{act} + \vbf_\text{act},
\end{equation}
where $\mubf_i^\text{act}$ is the mean of the Gaussian, and all vectors are expressed in the actor coordinate system. Further, we can express this velocity in the coordinate system of sensor $C$ using the composition of the actor-to-world and world-to-camera transforms, $T^{\text{act}\to C}$, as
\begin{equation}
    \vbf_{i, \text{dyn}}^{C} = T^{\text{act}\to C} \vbf_{i,\text{dyn}}^{\text{act}}.
\end{equation}

To account for both the velocity from the sensor and the velocity from the dynamic actor, we combine the two velocity sources. We add the velocity contributions together and obtain the complete velocity of Gaussian $i$ relative to sensor $C$ as
\begin{equation}
    \begin{split}
    \vbf_{i}^C &= \vbf_{i, \text{static}}^C + \vbf_{i, \text{dyn}}^{C} \\
    &= -(\omegabf_C \times \mubf_i^C + \vbf_C) + T^{\text{act}\to C} (\omegabf_\text{act} \times \mubf_i^\text{act} + \vbf_\text{act}).
    \end{split}
\end{equation}
Here, the transformed velocity from the dynamic actor retains its sign as it is already relative to the sensor.
Finally, using the derivation in \cite{seiskari2024gaussian} for the derivative of pixel coordinates with respect to camera motion, we obtain the pixel velocity for Gaussian $i$ projected into sensor $C$ by multiplying the result with the Jacobian of the projective transform
\begin{equation}
    \vbf_i^I = \Jbf^I (  -\omegabf_C \times \mubf_i^C - \vbf_C + T^{\text{act}\to C} (\omegabf_\text{act} \times \mubf_i^\text{act} + \vbf_\text{act}) ).
\end{equation}

\section{Additional qualitative results}
\label{appendix:results}
\begin{figure*}[t]
    \centering
    \includegraphics[width=0.95\linewidth]{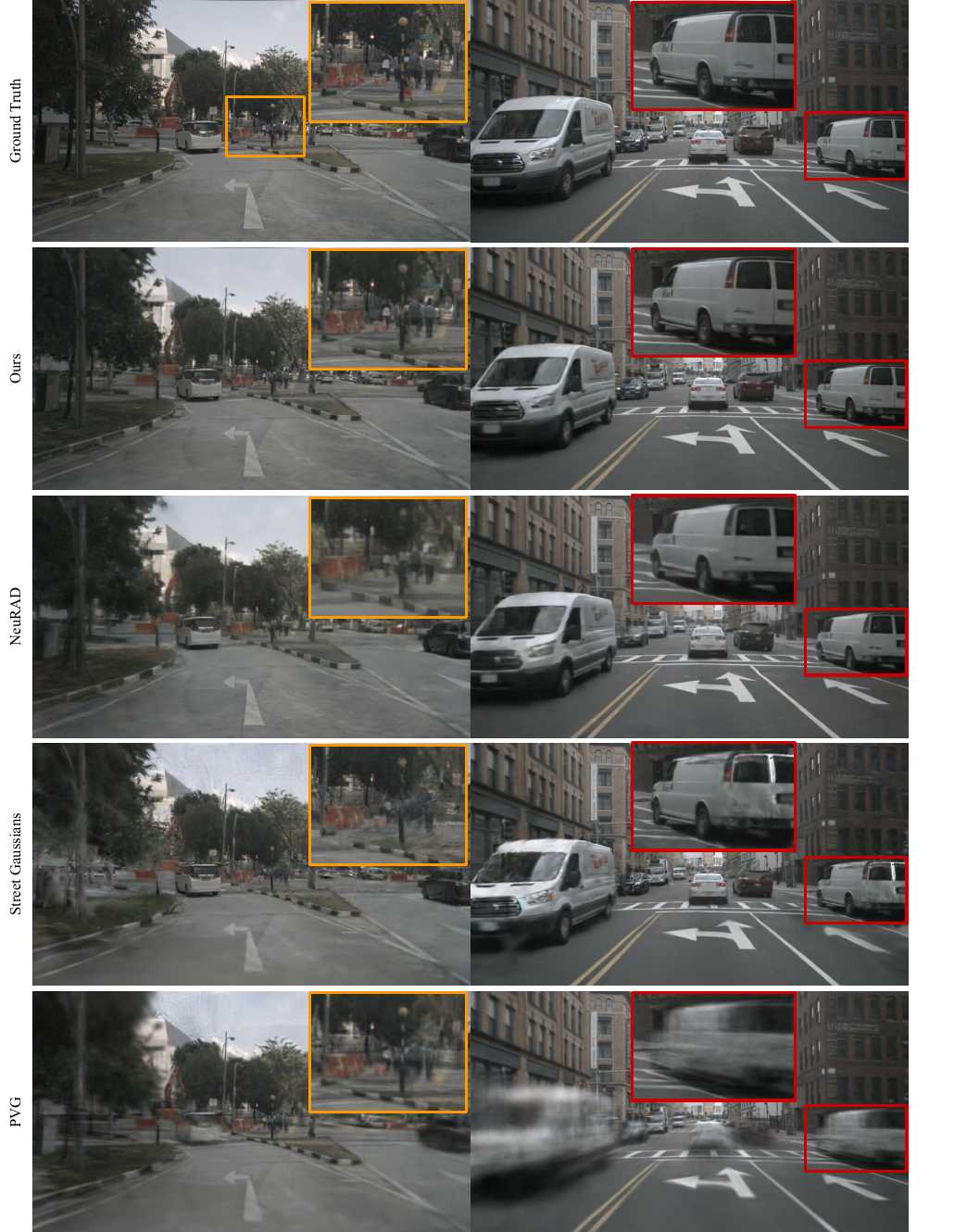}
    \caption{Qualitative NVS examples for nuScenes.}
    \label{fig:qualitative_examples_nuscenes}
\end{figure*}

\begin{figure*}[t]
    \centering
    \includegraphics[width=0.95\linewidth]{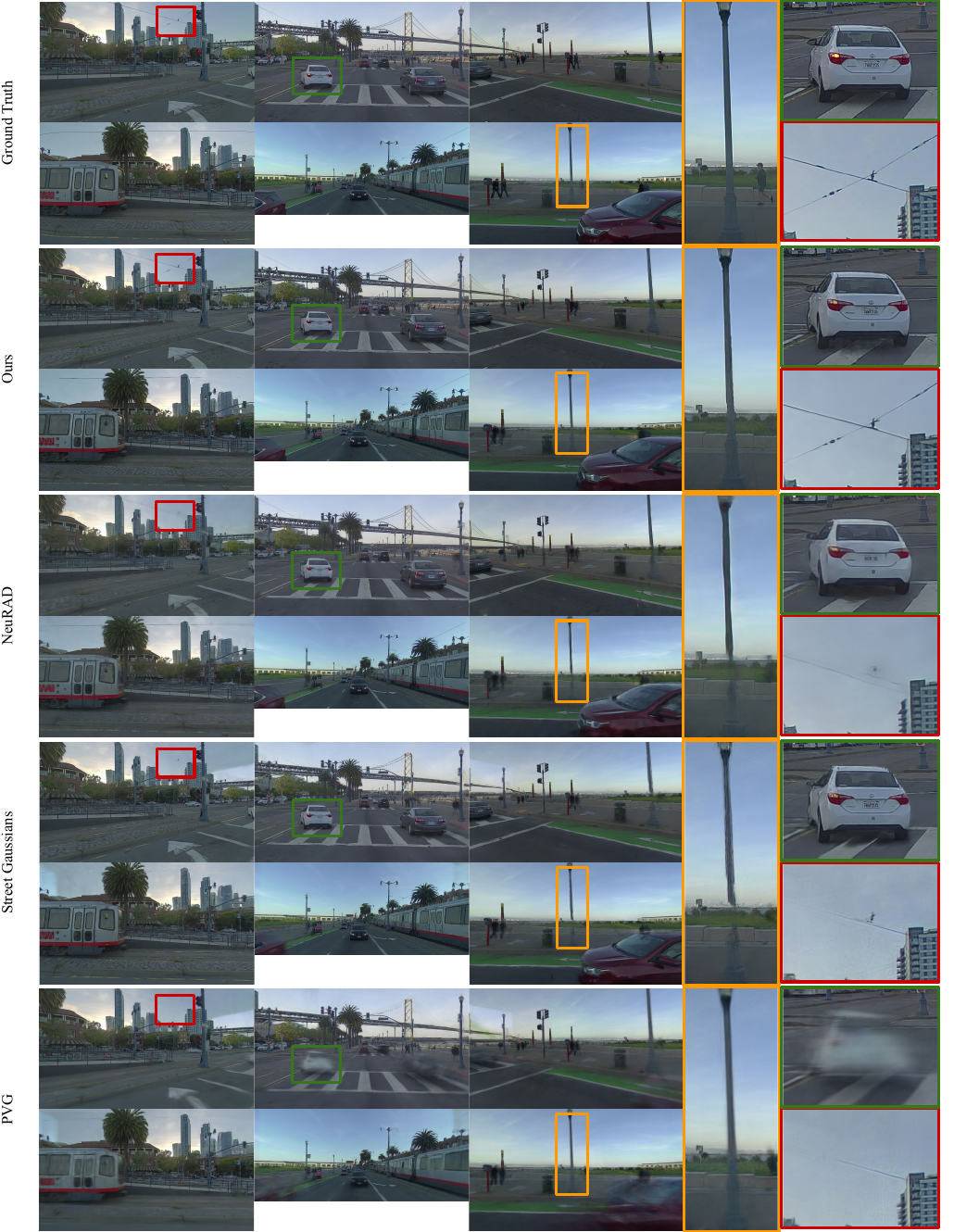}
    \caption{Qualitative NVS examples for PandaSet.}
    \label{fig:qualitative_examples_pandaset}
\end{figure*}

\begin{figure*}[t]
    \centering
    \includegraphics[width=0.95\linewidth]{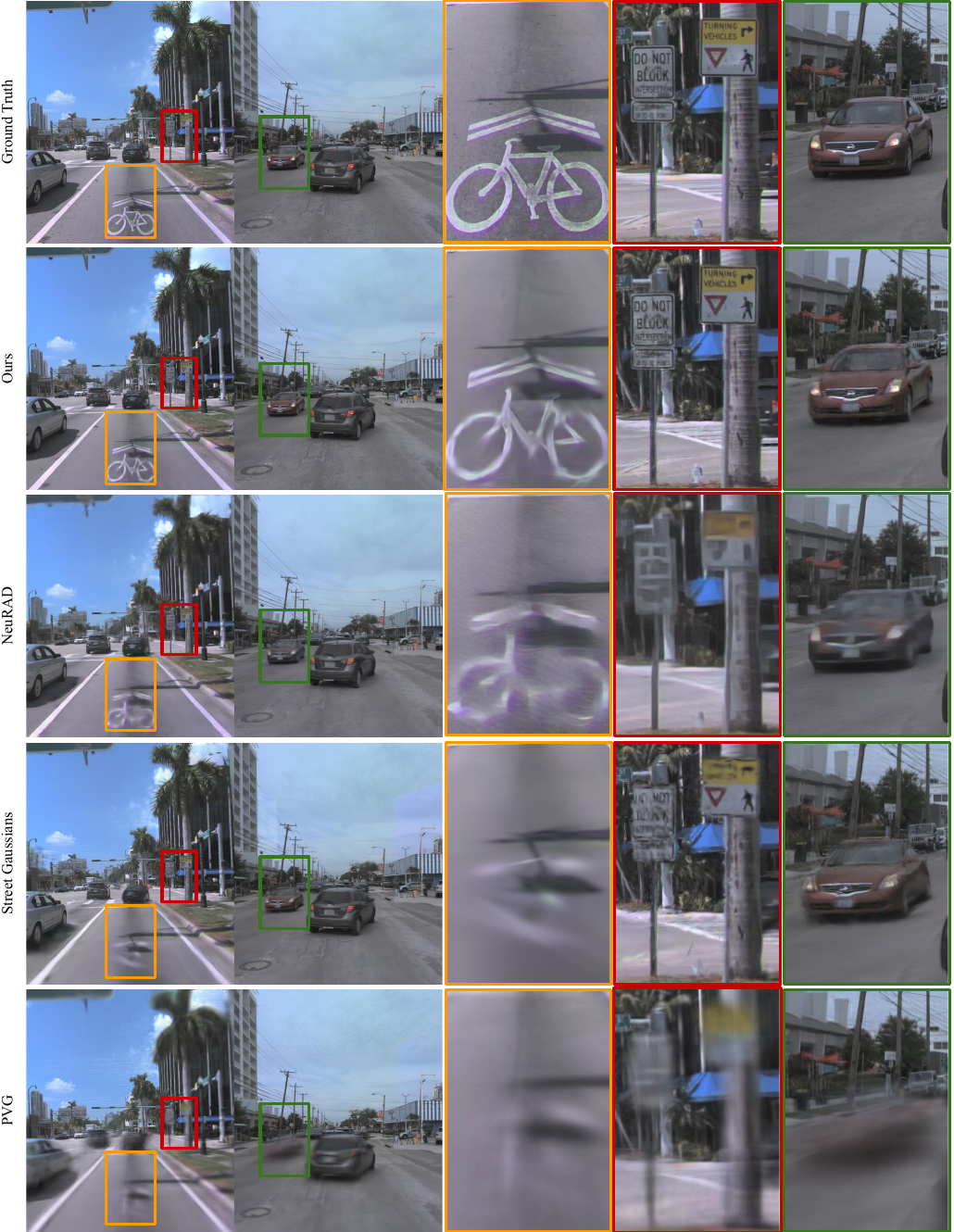}
    \caption{Qualitative NVS examples for Argoverse2.}
    \label{fig:qualitative_examples_argoverse2}
\end{figure*}

\begin{figure*}[t]
    \centering
    \includegraphics[width=0.9\linewidth,  trim={0cm 17.0cm 0cm 0cm},clip]{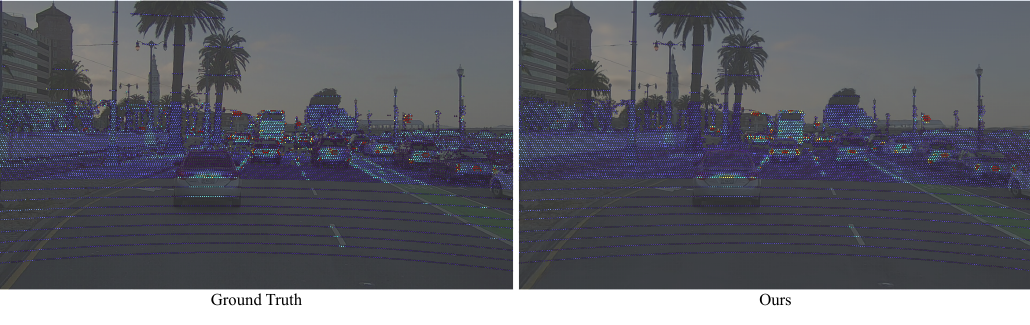}
    \caption{Qualitative NVS example for our lidar intensity rendering. We illustrate a rendered point cloud painted with predicted intensity, projected into and overlaid on the corresponding RGB image. The RGB image has been made darker and more transparent to highlight the intensity colors.}
    \label{fig:qualitative_examples_intensity}
\end{figure*}

We provide additional qualitative comparisons for the NVS task between SplatAD and our baselines for nuScenes (\cref{fig:qualitative_examples_nuscenes}), PandaSet (\cref{fig:qualitative_examples_pandaset}), and Argoverse 2 (\cref{fig:qualitative_examples_argoverse2}). We omit OmniRe from the comparisons, as its renderings closely resemble Street Gaussians in most cases. Further, we show a qualitative example of our lidar rendering (\cref{fig:qualitative_examples_intensity}), highlighting our method's ability to predict realistic intensity values.


\section{Evaluation details}
\label{appendix:eval_details}
Here, we present the dataset-specific details of our evaluation. The same evaluation protocol is used for all datasets. For the NVS task, we adopt a 50\% split, \ie, using every other frame for training and the remaining frames for hold-out validation. In the reconstruction task, we train and evaluate using all frames and lidar scans.

\parsection{PandaSet}
We use the complete sensor rig of six cameras and one lidar when training and evaluating on PandaSet. We crop out the bottom 260 pixels from the back camera to remove views of the ego-vehicle. We choose the same 10 sequences as in \cite{unisim} and \cite{tonderski2024neurad}: \texttt{001, 011, 016, 028, 053, 063, 084, 106, 123, 158}.

\parsection{Argoverse2}
For Argoverse2, we use the seven ring cameras and both lidars. We crop out the bottom 250 pixels of the front center, rear left, and rear right cameras to remove views of the ego-vehicle. Again, we choose the same 10 sequences as \cite{tonderski2024neurad}: \texttt{
05fa5048-f355-3274-b565-c0ddc547b315, 0b86f508-5df9-4a46-bc59-5b9536dbde9f, 185d3943-dd15-397a-8b2e-69cd86628fb7, 25e5c600-36fe-3245-9cc0-40ef91620c22, 27be7d34-ecb4-377b-8477-ccfd7cf4d0bc, 280269f9-6111-311d-b351-ce9f63f88c81, 2f2321d2-7912-3567-a789-25e46a145bda, 3bffdcff-c3a7-38b6-a0f2-64196d130958, 44adf4c4-6064-362f-94d3-323ed42cfda9, 5589de60-1727-3e3f-9423-33437fc5da4b
}.

\parsection{nuScenes}
We use all six available cameras and the top lidar on the nuScenes dataset. The bottom 80 pixels of the back camera are cropped to remove any views of the ego-vehicle. We select the following 10 sequences: \texttt{0039, 0054, 0061, 0066, 0104, 0108, 0122, 0176, 0180, 0193}.